\documentclass[letterpaper]{article} 
\usepackage{aaai2026}  
\usepackage{times}  
\usepackage{helvet}  
\usepackage{courier}  
\usepackage[hyphens]{url}  
\usepackage{graphicx} 
\usepackage{natbib}  
\usepackage{caption} 
\usepackage{mwe}
\usepackage{graphicx}
\usepackage{amsmath}
\usepackage{algorithm}
\usepackage{algpseudocode}
\usepackage{tabularray}
\usepackage{booktabs}

\usepackage{newfloat}
\usepackage{listings}
\DeclareCaptionStyle{ruled}{labelfont=normalfont,labelsep=colon,strut=off} 
\floatstyle{ruled}
\newfloat{listing}{tb}{lst}{}
\floatname{listing}{Listing}
\title{DFDT: Dynamic Fast Decision Tree for IoT Data Stream Mining on Edge Devices}
\author{
    Afonso Lourenço\textsuperscript{\rm 1}, João Rodrigo\textsuperscript{\rm 1}, João Gama\textsuperscript{\rm 2}, Goreti Marreiros\textsuperscript{\rm 1}}
\affiliations{
    \textsuperscript{\rm 1}GECAD, ISEP, Polytechnic of Porto, Rua Dr. António Bernardino de Almeida, Porto, 4249-015, Portugal\\
    \textsuperscript{\rm 2}INESC-TEC, FEP, University of Porto, Rua Dr. Roberto Frias, Porto, 4200-465, Portugal
}

\begin{document}

\maketitle

\begin{abstract}
The Internet of Things generates massive data streams, with edge computing emerging as a key enabler for online IoT applications and 5G networks. Edge solutions facilitate real-time machine learning inference, but also require continuous adaptation to concept drifts. While extensions of the Very Fast Decision Tree (VFDT) remain state-of-the-art for tabular stream mining, their unregulated growth limit efficiency, particularly in ensemble settings where post-pruning at the individual tree level is seldom applied. This paper presents DFDT, a novel memory-constrained algorithm for online learning. DFDT employs activity-aware pre-pruning, dynamically adjusting splitting criteria based on leaf node activity: low-activity nodes are deactivated to conserve resources, moderately active nodes split under stricter conditions, and highly active nodes leverage a skipping mechanism for accelerated growth. Additionally, adaptive grace periods and tie thresholds allow DFDT to modulate splitting decisions based on observed data variability, enhancing the accuracy–memory–runtime trade-off while minimizing the need for hyperparameter tuning. An ablation study reveals three DFDT variants suited to different resource profiles. Fully compatible with existing ensemble frameworks, DFDT provides a drop-in alternative to standard VFDT-based learners.
\end{abstract}

\section{Introduction}

The Internet of Things (IoT) connects a vast network of physical devices, generating massive, high-speed data streams \cite{gaber2014data}. To extract valuable insights from this ever-growing stream, the edge computing paradigm has emerged as a key enabler for low-latency IoT applications and 5G networks, bridging the gap between cloud services and end-users \cite{lourencco2025device}. However, benefiting from edge solutions not only implies bringing machine learning models for real-time inference, but also continuous updates to adapt to the evolving nature of unbounded streams. Unlike batch learning, where all training data is available upfront, data streams arrive incrementally \cite{gama2010knowledge,lourencco2025context}. Thus, making models susceptible to becoming outdated, especially due to concept drifts, i.e., changes in data distribution over time.

\begin{figure}[ht]
    \centering
    \includegraphics[width=1\linewidth]{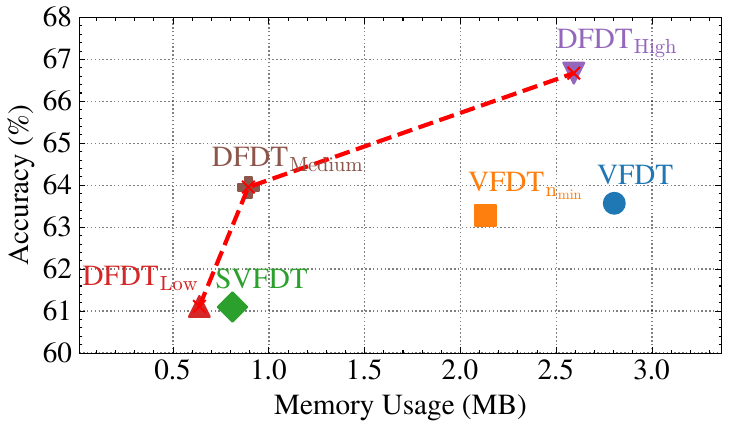}
    \caption{\textit{Accuracy} x \textit{Memory}}
    \label{fig:pareto_memory}
\end{figure}

To circumvent these challenges, many stream mining algorithms have been proposed, with extensions of the Very Fast Decision Tree (VFDT) \cite{domingos2000mining} being the state-of-the-art for tabular data sources. Their success can be attributed to approximation-based splitting, with incrementally updated statistical summaries of entropy-based metrics, e.g. information gain, to determine whether the observed utility of a split is statistically significant when the data distribution is unknown, e.g. via the Hoeffding bound (HB). This enables trees to grow incrementally, refining their partitions of the input space as more data arrives. However, uncontrolled tree growth can lead to excessive memory usage and reduced adaptability.

To counter this, adaptive tree algorithms introduce post-pruning mechanisms for forgetting. For instance, CVFDT \cite{hulten2001mining} evaluates alternate subtrees in parallel, replacing them when they outperform their predecessors. HAT \cite{bifet2009adaptive} generalizes this with multiple recursive replacements, while EFDT \cite{manapragada2018extremely} aggressively prunes subtrees when better local structures emerge. Other methods, like UFFT \cite{gama2005learning} and OnlineTree2 \cite{nunez2007learning}, use a short-term memory to stage and evaluate candidate subtrees, enabling smoother transitions and compact models.

Despite these innovations, single decision trees often suffer from instability and limited predictive performance. As a result, the community has largely shifted towards ensemble-based methods, where multiple diverse trees are combined to enhance robustness and adaptability \cite{gomes2017survey, krawczyk2017ensemble}. Ensembles handle concept drift at a higher level, dynamically updating individual components based on performance signals and using drift detectors to add, remove, or reweight learners \cite{neves2025online}. Thus, reducing the need for post-pruning at the individual tree level. As a result, most state-of-the-art ensembles still adopt the original VFDT as their base learner, often leading to inefficient memory use and redundant computation. Since post-pruning is rarely applied in ensemble settings, this raises a critical question: can pre-pruning strategies be leveraged to proactively regulate VFDT's growth, preventing unnecessary expansion in uncertain or low-utility regions of the input space?

To this end, a novel algorithm tailored for memory-constrained data stream mining is introduced, coined as DFDT: Dynamic Fast Decision Tree, whose main contributions are:

\begin{itemize}
    \item different splitting conditions according to leaf activity, so that the algorithm can grow more organically depending on the distribution and number of instances observed at each leaf of the tree:
    \begin{itemize}
        \item for nodes with low activity, DFDT deactivates the leaf node, conserving memory and computational resources.
        \item for nodes with moderate activity, DFDT applies conservative splitting constraints based on entropy, information gain, and the number of instances observed at the leaf, ensuring controlled growth.
        \item for highly active nodes, DFDT potentially employs a skipping mechanism, enabling rapid growth in response to significant data changes.
    \end{itemize}
    \item dynamically adjustable grace periods and tie thresholds, allowing the algorithm to either delay or accelerate splits depending on data variability. This flexibility improves the accuracy–memory–runtime trade-off and reduces the need for extensive hyperparameter tuning.
\end{itemize}

An ablation study of these DFDT’s components reveals three competitive variants tailored to different accuracy–memory trade-offs, summarized in Figure~\ref{fig:pareto_memory}. These variants are fully compatible with existing ensemble frameworks \cite{gomes2017survey, krawczyk2017ensemble}, offering a drop-in alternative to standard VFDT-based learners.

\section{Related Work}
\label{sec:related}

The core component for incrementally constructing decision trees (DTs) on edge devices, while maintaining a fixed time complexity per sample, is approximation-based splitting \cite{domingos2000mining}. As new instances arrive, they are processed from the root to a leaf node, updating statistics at each node. These updates are used to periodically adjust heuristic values for each attribute, such as information gain (IG) and Gini index (GI) \cite{domingos2000mining}. The tree attempts to perform splits based on a statistical bound that quantifies the confidence interval for the heuristic function, given a minimum amount of data. Typically, DTs compare the Hoeffding bound (HB) to the difference in evaluation between the best and second-best attribute splits \cite{domingos2000mining}. When this difference exceeds the bound, the leaf node is split into child nodes.

\paragraph{Splitting rules.} While traditional incremental DTs exclusively focus on evaluating the top two attributes, one can introduce extra splitting rules to promote even more valuable splits. For example, M-VFDT \cite{yang2011moderated} incorporates the fluctuation of the HB, tracking its mean, minimum, and maximum values, along with an accuracy metric, as a pre-pruning condition for split decisions. Alternatively, SVFDT \cite{da2018strict} applies constraints that compare current metrics against historical data and cross-leaf information, while introducing a skipping mechanism to bypass these constraints when significant changes in the data are detected. Furthermore, REG-VFDT \cite{barddal2020regularized} adds a constraint to assess whether the best-ranked feature, chosen for splitting at a leaf node, provides substantial gains compared to the gains observed in previous splits within the same branch.

\paragraph{Tie breaking threshold ($\tau$).} A critical aspect of incremental DTs is the tie-breaking procedure. While using the statistical difference in IG between two attributes helps control tree growth, competition between two equally favored split candidates can hinder progress, especially when either option would be equally suitable. To mitigate this, if the difference in heuristic values exceeds a predefined tie threshold, denoted as $\tau$, the split is performed. This threshold effectively controls the minimum rate of tree growth, with the attributes’ ability to separate before reaching the threshold influencing the speed at which the tree expands. However, a fixed $\tau$ value may cause ties to be broken prematurely, before a meaningful decision can be made, due to a lack of suitable candidates rather than a true tie situation. To address this, VFDT-$\tau$ \cite{holmes2005tie} compares the difference between the best and second-best candidates with the difference between the second-best and worst candidates, while M-VFDT \cite{yang2011moderated} designs an adaptive tie threshold that is dynamically calculated from the mean of HB, which has been shown to be proportionally related to the input stream samples.

\paragraph{Grace period ($n_{\min}$).} To prevent premature splits with small sample sizes that undermine the validity of the HB, a grace period \( n_{\min} \) specifies the minimum number of instances a leaf must observe before considering a split. However, without leveraging any information from the processed data, this can still result in computationally expensive split attempts or unnecessary delays in predictions. An alternative approach involves detecting frequent tie-breaking situations and applying steps to reduce the frequency of ties. For example, the default tie-breaking wait period can be adjusted by increasing the wait period for the child nodes after each tie is broken. This effect is cumulative until a valid split is found, after which the wait period is reset to the default value \cite{holmes2005tie}. Additionally, local statistics can be used to predict the optimal split time, minimizing delays and unnecessary split attempts. For instance, OSM \cite{losing2018enhancing} uses class distributions from previous split attempts to estimate the minimum number of examples required before the HB. Alternatively, VFDT-\(n_{\min}\) \cite{garcia2018hoeffding} adjusts the grace period after unsuccessful split attempts, according to the reason of failure in order to ensure a split in the next iteration. For example, if the best attributes are not too similar, but their IG difference is insufficient to trigger a split, the solution is to wait for additional examples until the HB decreases enough to be smaller than the IG difference, adjusting the \( n_{\min} \) accordingly. Similarly, if the top attributes are very similar in terms of IG, but the tie threshold \( \tau \) is not exceeded, more instances are needed to allow the HB to decrease below \( \tau \), with \( n_{\min} \) adjusted based on their IG difference \cite{garcia2018hoeffding}.

\paragraph{Leaf activity.} Another memory-efficient strategy is to condition split decisions based on leaf activity. GAHT \cite{garcia2022green} devises a normalized measure to quantify how many instances have been observed at a particular leaf relative to the mean number of instances per leaf since its creation. Based on this value, nodes can either be deactivated when their activity is low, halting further splits to conserve resources, or more aggressive expansion strategies can be applied when activity exceeds a threshold, accelerating the growth of important nodes by relaxing the splitting rule to a less strict alternative. These adaptive expansion modes offer a more nuanced approach to tree growth, ensuring computational resources are focused on expanding nodes that significantly influence the model’s accuracy, while avoiding unnecessary splits in less relevant parts of the tree. Alternatively, VFDT \cite{domingos2000mining} deactivates the least promising leaves based on the probability that examples will reach those leaves and their observed error rate. If a better split is found, the leaves of that split are deactivated, and their statistics are preserved. A new split is then performed, creating new leaves. If, during split re-evaluation, a previously deactivated split is found to be the best option, the saved statistics are restored rather than starting the process from scratch.

\textbf{Overview.} Table~\ref{table:optimizations} summarizes how various algorithms implement different pre-pruning strategies, categorized by whether they adapt rules, grace periods, tie thresholds, or activity-based controls. As shown, DFDT offers the most comprehensive integration of these mechanisms, seamlessly combining them in a cohesive manner, as will be discussed.

\begin{table}[ht]
\centering
\caption{Incremental decision trees}
\label{table:optimizations}
\begin{tabular}{ccccccc}
\toprule
\textbf{Year} & \textbf{Model} & \textbf{Activity} & \textbf{Rules} & \textbf{\text{$n_{\text{min}}$}} & \textbf{$\tau$} \\
\midrule
2000 & VFDT & X & &  &  \\
2005 & VFDT-$\tau$ &  & X &  & X \\
2011 & M-VFDT  &  & X &  & X \\
2018 & SVFDT & & X & & \\
2018 & OSM & & & X & \\
2018 & VFDT-$n_{\text{min}}$ & & & X &\\
2020 & REG-VFDT & & X &  &  \\
2022 & GAHT & X & &  & \\
2026 & DFDT & X & X & X & X \\
\bottomrule
\end{tabular}
\end{table}

\section{Methodology}
\label{sec:algorithm}

DFDT modifications are here described in detail, while referring to the pseudo-code in Algorithm \ref{alg}. The initialization phase creates the root node structure of the decision tree (Alg. 1, Step 2, 3), sets up estimators (Alg. 1, Step 4), and instance counters (Alg. 1, Step 5). This initialization involves only constant-time operations, resulting in a computational complexity of \(O(1)\), independent of the number of instances or features. The main loop iterates over each instance \((X, y)\) in the data stream \(S\), running \(N\) times in total. For each instance, several steps are performed. First, the instance is routed through the tree to the corresponding leaf node, where the prediction \(\hat{y}\) is obtained (Alg. 1, Step 7). Assuming a balanced tree, the depth of traversal is \(O(\log_B |LH|)\), where \(|LH|\) is the number of leaf nodes and \(B\) is the branching factor. The prediction at the leaf node is a constant-time operation, \(O(1)\). Following this, the instance count and feature estimators are updated at the leaf node (Alg. 1, Steps 8-10). This update takes \(O(F)\) time.

\begin{algorithm}[h!]
\caption{Dynamic Fast Decision Tree (DFDT)}
\label{alg}
\begin{algorithmic}[1]
    \Procedure{DFDT}{$S$: data stream, $\delta$: confidence level, $f_{\text{expand}}$: threshold for skipping, $f_{\text{deactivate}}$: threshold to deactivate leaves}
        \State $DFDT \leftarrow$ root
        \State $LH \leftarrow \text{hash of leaves}$ 
        \State Initialize $H_{LH_{\text{stat}}}, n_{\text{stat}}, H_{\text{stat}}, G_{\text{stat}}, HB_{\text{stat}}, n_{\text{min}}$
        \State $n, n_{\text{l}}, n_{\text{check}_{\text{l}}}, n_{\text{leaf}_{\text{l}}}, n_{\text{tree}_{\text{l}}} \leftarrow 0$ where $l=\text{root}$
        \For{each $(X, y)$ in $S$}
            \State Route $(X, y)$ to leaf $l$ and obtain prediction $\hat{y}$
            \State Update $l$ feature estimators and class distribution 
            \State $n  \leftarrow n + 1$
            \State $n_l  \leftarrow n_l + 1$
            \If{$\frac{(n_l - n_{\text{leaf}_l}) \times |LH|}{n-n_{\text{tree}_{l}}} < f_{\text{deactivate}}$}
                \State Deactivate $l$
            \ElsIf{$> f_{\text{expand}}$}
                \State GrowFast $\leftarrow$ True
            \EndIf
            \If{(impure at $l$) and ($n_l - n_{\text{check}_l} > n_{\text{min}}$)}
                \State Update $HB_{stat}$ with $\epsilon$
                \If{\textsc{CanSplit}(GrowFast)}
                    \State Replace leaf $l$ with a split node
                    \For{each leaf branch $b$ of the split}
                        \State Initialize estimators
                        \State Set post-split distribution at $b$
                        \State Update $LH$
                        \State $n_b, n_{\text{leaf}_b}, n_{\text{check}_b} \leftarrow \sum_{c \in \mathcal{C}} n_{b}^{(c)}$
                        \State $n_{\text{tree}_{b}} \leftarrow  n$
                    \EndFor
                \Else
                    \State $n_{\text{check}_l} \leftarrow n_l$
                    \If{$\overline{HB}_{\text{stat}} < \Delta G < \epsilon$}
                        \State $n_{\text{min}} \leftarrow \left\lceil \frac{R^2 \ln(1/\delta)}{2 (\Delta G)^2} \right\rceil$
                    \ElsIf{$\Delta G < \overline{HB}_{\text{stat}} < \epsilon$}
                        \State $n_{\text{min}} \leftarrow \left\lceil \frac{R^2 \ln(1/\delta)}{2 (\overline{HB}_{\text{stat}})^2} \right\rceil$
                    \EndIf
                \EndIf
            \EndIf
        \EndFor
    \EndProcedure
\end{algorithmic}
\end{algorithm}

While the HB ensures that splits are made with statistical confidence, it treats all nodes equally in terms of expansion, despite the fact that not all nodes contribute equally to the decision-making process. To improve VFDT's efficiency in allocating computational resources, a notion of relative importance can be introduced to dynamically adjust the rate of expansion at each node. To assess the activity level at each node, a fraction parameter is calculated as:

\begin{equation}
\text{f} = \frac{(n_l - n_{\text{leaf}_l}) \times |LH|}{n-n_{\text{tree}_{l}}}
\end{equation}

where $(n_l - n_{\text{leaf}_l})$ is the number of instances observed at a particular leaf $l$ since its creation, $n-n_{\text{tree}_{l}}$ the total instances observed by the tree since the creation of leaf $l$, and $|LH|$ the current total number of leaves. If the fraction is below a threshold $f_{\text{deactivate}}$, the leaf node is deactivated, halting further splits to save on resources (Alg. 1, Steps 12), which is a constant-time operation, \(O(1)\). If the metric falls above a threshold $f_{\text{expand}}$, a boolean saves the intention to apply a more aggressive expansion strategy, accelerating the growth of important nodes (Alg. 1, Step 14).

To avoid costly evaluations, the algorithm only checks for potential splits once a leaf node accumulates \(n_{\text{min}}\) instances since the last splitting opportunity (Alg. 1, Step 16). Then a split attempt is performed (Alg. 1, Step 18), with its pseudocode detailed in Algorithm \ref{alg2}. This check involves computing \(G(\cdot)\) values, requiring \(O(F)\) time, and sorting the \(G(\cdot)\) values (Alg. 2, Step 2), yielding a time complexity of \(O(F \log F)\). Next, the algorithm verifies whether the HB or tie-breaking threshold is satisfied (Alg. 2, Step 4). While in the VFDT, the tie threshold (\(\tau\)), which effectively controls minimum growth speed, remains static, DFDT, instead, dynamically sets the tie threshold as the mean of the Hoeffding Bound values computed over the most recent
\(k\) instances \cite{yang2011moderated}. This reduces the need for trial-and-error in selecting a fixed threshold, optimizing tree performance in dynamic environments. As the data stream becomes noisier, the adaptive \(\tau\) stabilizes the model. If the HB is satisfied, DFDT utilizes adaptive expansion modes, determined by the fraction-determined boolean. For nodes with moderate activity, DFDT applies four conservative splitting constraints \cite{da2018strict}, ensuring controlled growth (Alg. 2, Step 12-15). For highly active nodes, i.e., when the fraction value exceeds 2, DFDT also checks two skipping conditions (Alg. 2, Step 5-11).

\begin{algorithm}[htbp]
\caption{Split conditions}
\label{alg2}
\begin{algorithmic}[1]
    \Procedure{CanSplit}{GrowFast = False}, 
            \State  \( G_{\text{best}} = \max(Sorted G(\cdot)) \)
            \State \( G_{\text{second best}} = \max(Sorted G(\cdot) \setminus \{G_{\text{best}}\}) \)
            \If{$G_{\text{best}} - G_{\text{second best}} \geq \epsilon$ or $\epsilon < \overline{HB}_{\text{stat}}$}
                \If{GrowFast}
                    \State $C1 \leftarrow H_l \geq \overline{H}_{\text{stat}} + \sigma(H_{\text{stat}})$ 
                    \State $C2 \leftarrow G_{\text{best}} \geq \overline{G}_{\text{stat}} + \sigma(G_{\text{stat}})$
                    \If{C1 $and$ C2} 
                        \State \Return True
                    \EndIf
                \EndIf
                \State $C3 \leftarrow H_l \geq \overline{H}_{LH_{\text{stat}}} - \sigma(H_{LH_{\text{stat}}})$
                \State $C4 \leftarrow H_l \geq \overline{H}_{\text{stat}} - \sigma(H_{\text{stat}})$
                \State $C5 \leftarrow G_{\text{best}} \geq \overline{G}_{\text{stat}} - \sigma(G_{\text{stat}})$
                \State $C6 \leftarrow n_l \geq \overline{n}_{\text{stat}}$
                \State Update $H_{stat}$ with $H_l$
                \State Update $G_{stat}$ with $G_{best}$
                \State Update $n_{stat}$ with $n_l$
                \If{C3 $and$ C4 $and$ C5 $and$ C6} 
                    \State \Return True
                \EndIf
            \EndIf
        \State \Return False
    \EndProcedure
\end{algorithmic}
\end{algorithm}

The four primary constraints governing the split operate on the instance count, global entropy, historical entropy, and historical information gain. The instance count constraint ensures that splits only occur when the leaf has accumulated sufficient data, by checking if the current instance count \( n_l \) at the leaf $l$ meets or exceeds the mean instance count recorded across all leaves under VFDT-satisfied conditions. Conversely, the remaining three constraints are evaluated using the same general formula \( \varphi(x, X) \), which determines that the current metric value \( x \) at a leaf is significantly different from reference values \( X \) when bigger than the mean of the values in \( X \) minus a standard deviation:

\begin{equation}
\varphi(x, X) = 
\begin{cases} 
   \text{True, if } x \geq \bar{X} - \sigma(X) \\
   \text{False, otherwise}
\end{cases}
\end{equation}

The global entropy constraint checks if the current entropy \( H_l \) of a leaf $l$ is high compared to the overall entropy of all leaves in the tree, denoted as $H_{LH_{\text{stat}}}$. Here, the set \( X \) consists of entropy values from all leaves, recorded continuously across the tree. The historical entropy constraint examines whether the current entropy \( H_l \) at a leaf $l$ is significantly higher than historical entropy values recorded across all leaves at the times when VFDT conditions were met, denoted as $H_{\text{stat}}$. This ensures that splits are performed only if the leaf’s entropy has increased meaningfully over time. The historical information gain constraint assesses whether the current information gain \( G_l \) of the best splitting feature at a leaf $l$ is notably higher than historical information gain values recorded across all leaves when VFDT conditions were met, denoted as $G_{\text{stat}}$. This avoids low-value splits.

Together, these constraints allow DFDT to restrict tree growth in nodes of moderate activity to instances with statistically significant differences. For highly active nodes, two skipping conditions are calculated according to \( \omega(x, X) \), which permits a split when the current metric \( x \), i.e. the entropy \( H_l \) and information gain \( G_{best} \), is at least one standard deviation above the mean historical values in \( X \):

\begin{equation}
    \omega(x, X) = 
    \begin{cases} 
       \text{True, if } x \geq \bar{X} + \sigma(X) \\
       \text{False, otherwise}
    \end{cases}
\end{equation}

These splitting conditions work as a skipping mechanism to the conservative constraints. In these cases, splits are allowed to proceed directly if the current entropy and information gain are significantly higher than historical means. Subsequently, the statistics for entropy, information gain, HB, and the number of instances seen are updated (Alg. 2, Step 16-18). All of these are constant-time operations. If the conditions are satisfied, a split is triggered (Alg. 2, Step 18). This split involves navigating the tree, sorting features, and creating new nodes, with the added complexity of re-initializing feature estimators for each branch (Alg. 1, Step 19-25). The time complexity becomes \(O(F \log F + B \cdot F)\), where \(B \cdot F\) represents the initialization cost for the new branches. If the split attempt fails at any point (Alg. 2, Step 23), the algorithm resets the instance count of the last checked split (Alg. 1, Step 28), and recalculates the grace period \( n_{\text{min}} \) based on specific conditions at each leaf (Alg. 1, Steps 29-32). This decision is based on the tie threshold \( \tau \), the difference in information gain between the top attributes \( \Delta G \) and the statistical confidence bound \( \epsilon \), derived from the HB \cite{garcia2018hoeffding}, with two possible scenarios. In the first scenario, \( \Delta G \) is smaller than \( \epsilon \), but still exceeds \( \tau \), suggesting that more data is needed to confirm the observed gain. In this case, \( n_{\text{min}} \) is increased to allow additional data accumulation, ensuring that the gain difference becomes statistically significant in the subsequent evaluation. In the second scenario, \( \Delta G \) is below  \( \tau \), indicating that the attributes are statistically similar, but \( \epsilon \) is still above \( \tau \). Here, \( n_{\text{min}} \) is further increased to delay the split until more data is gathered, allowing \( \epsilon \) to shrink sufficiently. Thus:

\begin{equation}
n_{\text{min}} = 
\begin{cases}
\left\lceil \frac{R^2 \ln(1/\delta)}{2 (\Delta G)^2} \right\rceil, & \text{if } \tau < \Delta G < \epsilon \\\\
\left\lceil \frac{R^2 \ln(1/\delta)}{2 \tau^2} \right\rceil, & \text{if } \Delta G < \tau < \epsilon
\end{cases}
\end{equation}

where \( \delta \) is a user-defined confidence parameter, and \( R \) is the range of heuristic values. This adjustment minimizes unnecessary computations and improves energy efficiency, while maintaining the model’s accuracy. Since split attempts (Alg. 1, Steps 16–35) occur only once every \(n_{\text{min}}\) instances, their overall contribution to the time complexity is weighted by the grace period factor. Thus, the total time complexity is: $N \cdot O(\log_B |LH| + F) + \frac{N}{n_{\text{min}}} \cdot O(F \log F + B \cdot F)$.

\section{Experiments}
\label{sec:experiments}

All models and algorithm components were implemented within the pystream library\footnote{https://github.com/vturrisi/pystream}, a Cython-based tool for data stream mining, reducing runtime overhead by compiling Python code into C. For evaluation, a diverse set of real-world datasets obtained from the USP Data Stream Repository\footnote{https://sites.google.com/view/uspdsrepository} were used \cite{Souza2020}. These are shown in Table \ref{tab:datasets}, encompassing both binary and multiclass classification tasks. The experiments were conducted using a prequential (test-then-train) evaluation strategy \cite{gama2009issues}, monitoring the interrelationships among \textit{accuracy}, \textit{memory} usage, and \textit{runtime}. All algorithms were optimized for each dataset using a grid search strategy, with grace periods ($n_{\text{min}}$) set to 100, 400, and 1000, and tie-splitting thresholds (\( \tau \)) configured to 0.01, 0.05, and 0.1 as the hyperparameter settings. The Hoeffding bound confidence level was fixed at $1 \times 10^{-7}$ across all experiments. DFDT’s adaptive control of grace periods and tie thresholds removes the need for tuning these specific hyperparameters, as they govern tree growth and thus become implicitly determined by the HB confidence. However, DFDT introduces two fixed activity thresholds. While this raises questions about their adaptivity, it is important to note that, in contrast to $n_{\text{min}}$ and \( \tau \), adapting these activity thresholds through the HB confidence is inappropriate, because activity operates conditionally on HB-driven splits and modulates leaf deactivation in ways that counteract tree growth. In practice, consistent with findings from GAHT \cite{garcia2022green}, we observed that a leaf-deactivation threshold below 2\% of the expected instance count ($f_{\text{deactivate}} = 0.02$) and a skip-check threshold at twice the expected count ($f_{\text{expand}} = 2$) provide robust performance across datasets. Majority class prediction was used at the leaf level for all models. 

\begin{table}[ht]
\centering
\begin{tabular}{lccc}
\toprule
\textbf{Name}  & \textbf{Classes} & \textbf{Features} & \textbf{Examples} \\ \midrule
OZONE        & 2              & 72              & 2,534           \\ 
NOAA     & 2              & 8               & 18,159          \\ 
METER       & 10             & 96              & 22,950          \\ 
ELEC      & 2              & 8               & 45,312          \\ 
RIALTO        & 10             & 27              & 82,250          \\ 
POSTURE         & 11             & 3               & 164,860         \\ 
KDDCUP99    & 23             & 41              & 494,021         \\ 
COVER     & 7              & 54              & 581,012         \\ 
POKER       & 10             & 11              & 829,201       \\ \bottomrule
\end{tabular}
\caption{Experimental datasets}
\label{tab:datasets}
\end{table}

\begin{figure*}[ht]
    \centering
    \includegraphics[width=1\linewidth]{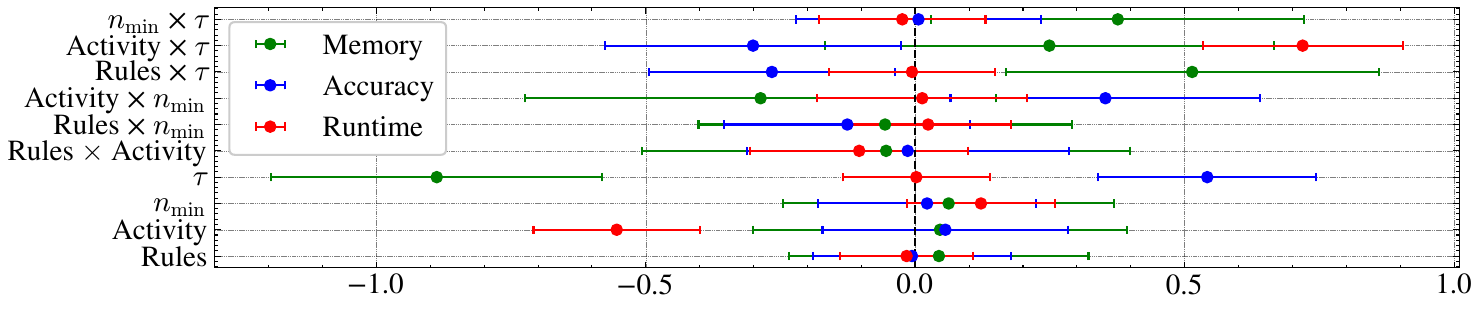}
    \caption{Main and interaction effects of DFDT components}
    \label{fig:two_way}
\end{figure*}

Since DFDT integrates multiple heuristics (Rules, Activity, $\tau$, and $n_{\text{min}}$), it is essential to conduct an ablation study to assess the individual and combined contributions of these components to runtime, memory usage, and accuracy. This analysis helps identify Pareto-efficient variants of the algorithm. To ensure fair comparison across datasets of varying scales, all results were normalized on a per-dataset basis. For interpretability, normalized memory and runtime values were inverted, so that higher scores consistently reflect better performance across all metrics. Figure~\ref{fig:two_way} presents coefficient estimates from linear regression models, each fitted separately for the three target outcomes. Interaction terms were included to capture the synergistic effects between heuristics. Individually, the adaptive grace period, activity-aware pruning, and stricter splitting rules did not produce statistically significant accuracy–memory–runtime trade-offs. Moreover, the combinations of Rules × Activity and Rules × $n_{\text{min}}$ also demonstrated stable trade-offs, informing the design of $\text{DFDT}_{\text{Low}}$ and $\text{DFDT}_{\text{Medium}}$, respectively. In contrast, the interaction between Activity × $n_{\text{min}}$ and the inclusion of an adaptive $\tau$ significantly improved accuracy, albeit at the expense of memory efficiency. Additionally, $\tau$'s interactions with other heuristics exhibited inconsistent effects on accuracy, suggesting a more complex dynamic. As a result, $\text{DFDT}_{\text{High}}$ activates all heuristics to maximize accuracy, potentially at the cost of runtime or memory consumption.

\begin{table}[ht]
\centering
\begin{tabular}{ccccc}
\toprule
\textbf{Model} & \textbf{Rules} & \textbf{Activity} & \textbf{\text{$n_{\text{min}}$}} & \textbf{$\tau$} \\
\midrule
$\text{DFDT}_{\text{\text{Low}}}$ & X & X &  &  \\
$\text{DFDT}_{\text{\text{Medium}}}$ & X &  & X & \\
$\text{DFDT}_{\text{\text{High}}}$ & X & X & X & X \\
\bottomrule
\end{tabular}
\caption{Pareto-efficient DFDT variants}
\label{table:proposed}
\end{table}

\begin{figure}[ht]
    \centering
    \includegraphics[width=1\linewidth]{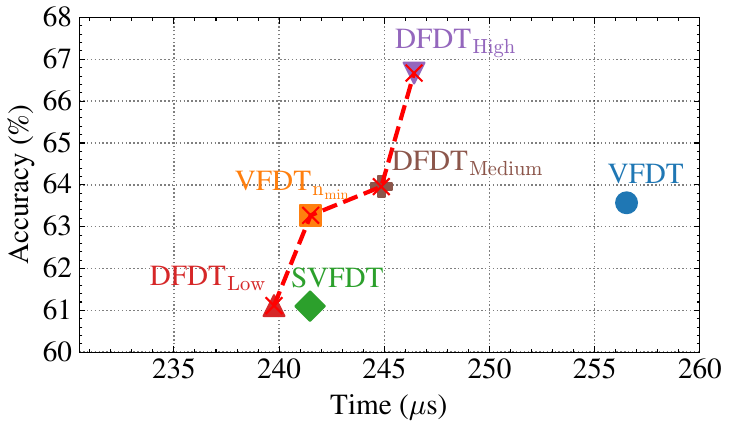}
    \caption{\textit{Accuracy} x \textit{Runtime}}
    \label{fig:pareto_time}
\end{figure}

Figures~\ref{fig:pareto_memory} and~\ref{fig:pareto_time} show the average performance across all datasets, plotting prequential accuracy against worst-case memory usage (MB) and per-instance computational time (µs), respectively. Table~\ref{tab:merged_results} reports the detailed results for each dataset. Among the baseline methods, VFDT exhibits moderate predictive performance while being the slowest and most memory-intensive model, underscoring its inefficiency across both metrics. Its modified variant, VFDT-$n_{\text{min}}$, offers a modest reduction in memory consumption and a notable improvement in runtime, with minimal impact on accuracy. SVFDT demonstrates the lowest memory usage and maintains competitive runtime with VFDT-$n_{\text{min}}$, however this efficiency is achieved at the expense of predictive accuracy. In contrast, the proposed DFDT variants delineate a clear Pareto frontier with respect to memory usage and share a competitive frontier with VFDT-$n_{\text{min}}$ in terms of runtime. Specifically, $\text{DFDT}_{\text{Low}}$ stands out as the most resource-efficient model, exhibiting lower runtime and memory consumption than SVFDT while maintaining comparable accuracy, making it particularly suitable for deployment in resource-constrained environments. The $\text{DFDT}_{\text{Medium}}$ variant achieves a favorable trade-off, significantly improving accuracy while preserving strong efficiency in both runtime and memory. Finally, $\text{DFDT}_{\text{High}}$ attains the highest overall accuracy, accompanied by a modest increase in computational time and a more pronounced increase in memory usage. Thus, only appropriate when resource constraints are less critical.

\begin{figure}[ht]
    \centering
    \includegraphics[width=1\linewidth]{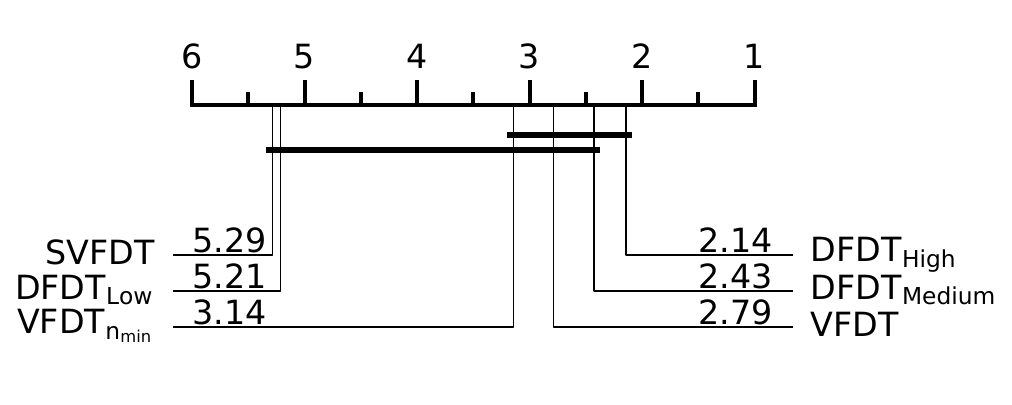}
    \caption{Nemenyi test - \textit{Accuracy}}
    \label{fig:accuracy}
\end{figure}

\begin{figure}[ht]
    \centering
    \includegraphics[width=1\linewidth]{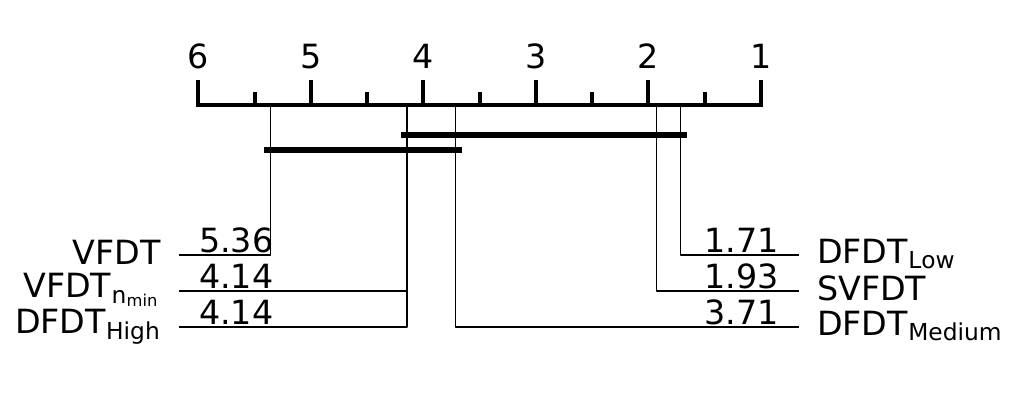}
    \caption{Nemenyi test - \textit{Memory}}
    \label{fig:memory}
\end{figure}

\begin{figure}[ht]
    \centering
    \includegraphics[width=1\linewidth]{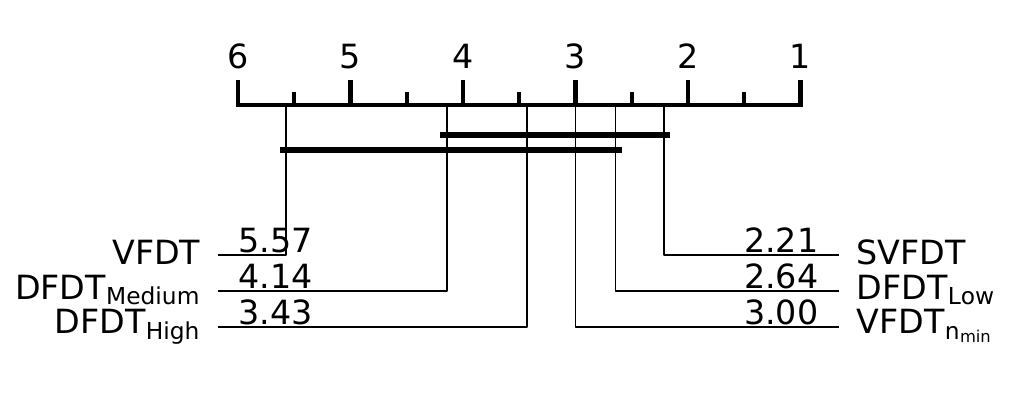}
    \caption{Nemenyi test - \textit{Runtime}}
    \label{fig:time}
\end{figure}

To assess the statistical significance of the observed differences, we applied the Friedman test followed by the Nemenyi post-hoc test. The resulting Critical Difference Diagrams (CDDs), shown in Figures~\ref{fig:accuracy}, \ref{fig:memory}, and~\ref{fig:time}, display the algorithm rankings at a 95\% confidence level. In these diagrams, lower values correspond to better average ranking across datasets. The CDD for accuracy reveals that $\text{DFDT}_{\text{High}}$ achieves the best overall rank, and is the only method that significantly outperforms SVFDT. This result confirms the effectiveness of incorporating all heuristic components within DFDT to maximize predictive performance.

\begin{table*}[ht]
\centering
\begin{tabular}{p{1.5cm}ccccccccc}
\toprule
\textbf{Method} &
\textbf{NOAA} &
\textbf{METER} &
\textbf{ELEC} &
\textbf{RIALTO} &
\textbf{POSTURE} &
\textbf{COVER} &
\textbf{POKER} &
\textbf{Avg. (Rank)} \\
\midrule

VFDT &
\begin{tabular}{@{}c@{}}73.1 $\pm$ 2.2\\0.47 MB\\43.2 µs\end{tabular} &
\begin{tabular}{@{}c@{}}52.4 $\pm$ 2.3\\14.19 MB\\288.5 µs\end{tabular} &
\begin{tabular}{@{}c@{}}\textbf{80.3 $\pm$ 1.5}\\0.14 MB\\49.8 µs\end{tabular} &
\begin{tabular}{@{}c@{}}35.0 $\pm$ 4.9\\3.37 MB\\118.1 µs\end{tabular} &
\begin{tabular}{@{}c@{}}50.5 $\pm$ 2.1\\0.22 MB\\78.5 µs\end{tabular} &
\begin{tabular}{@{}c@{}}79.5 $\pm$ 2.5\\0.80 MB\\254.1 µs\end{tabular} &
\begin{tabular}{@{}c@{}}74.2 $\pm$ 4.5\\0.43 MB\\963.4 µs\end{tabular} &
\begin{tabular}{@{}c@{}}63.6 (2.79)\\2.80 (5.36)\\256.5 (5.57) \end{tabular} \\

\midrule

VFDT$_{n_{\text{min}}}$ &
\begin{tabular}{@{}c@{}}72.6 $\pm$ 2.0\\0.31 MB\\35.0 µs\end{tabular} &
\begin{tabular}{@{}c@{}}52.4 $\pm$ 2.1\\10.37 MB\\285.1 µs\end{tabular} &
\begin{tabular}{@{}c@{}}79.4 $\pm$ 1.6\\0.10 MB\\38.4 µs\end{tabular} &
\begin{tabular}{@{}c@{}}34.1 $\pm$ 4.3\\2.87 MB\\ \textbf{104.7 µs} \end{tabular} &
\begin{tabular}{@{}c@{}}50.7 $\pm$ 1.5\\0.21 MB\\69.3 µs\end{tabular} &
\begin{tabular}{@{}c@{}}78.3 $\pm$ 2.4\\0.62 MB\\241.2 µs\end{tabular} &
\begin{tabular}{@{}c@{}}75.4 $\pm$ 4.5\\0.43 MB\\916.8 µs\end{tabular} &
\begin{tabular}{@{}c@{}}63.3 (3.14)\\2.13 (4.14)\\241.5 (3.00)\end{tabular} \\

\midrule

SVFDT &
\begin{tabular}{@{}c@{}}71.2 $\pm$ 2.0\\\textbf{0.15 MB}\\\textbf{33.0 µs}\end{tabular} &
\begin{tabular}{@{}c@{}}52.3 $\pm$ 1.9\\3.73 MB\\283.0 µs\end{tabular} &
\begin{tabular}{@{}c@{}}79.0 $\pm$ 0.6\\\textbf{0.05 MB}\\\textbf{35.7 µs}\end{tabular} &
\begin{tabular}{@{}c@{}}32.1 $\pm$ 3.4\\0.87 MB\\109.0 µs\end{tabular} &
\begin{tabular}{@{}c@{}}48.8 $\pm$ 1.7\\0.10 MB\\\textbf{69.0 µs}\end{tabular} &
\begin{tabular}{@{}c@{}}74.6 $\pm$ 2.0\\\textbf{0.39 MB}\\255.3 µs\end{tabular} &
\begin{tabular}{@{}c@{}}69.7 $\pm$ 2.0\\0.38 MB\\\textbf{905.3 µs}\end{tabular} &
\begin{tabular}{@{}c@{}}61.1 (5.29)\\0.81 (1.93)\\ 241.5 (2.21)\end{tabular} \\

\midrule

$\text{DFDT}_{\text{Low}}$ &
\begin{tabular}{@{}c@{}}72.4 $\pm$ 1.8\\0.16 MB\\33.3 µs\end{tabular} &
\begin{tabular}{@{}c@{}}52.3 $\pm$ 1.9\\\textbf{2.76 MB}\\279.1 µs\end{tabular} &
\begin{tabular}{@{}c@{}}79.1 $\pm$ 0.9\\0.07 MB\\37.8 µs\end{tabular} &
\begin{tabular}{@{}c@{}}33.9 $\pm$ 4.1\\\textbf{0.54 MB}\\110.2 µs\end{tabular} &
\begin{tabular}{@{}c@{}}48.4 $\pm$ 1.1\\\textbf{0.09 MB}\\70.3 µs\end{tabular} &
\begin{tabular}{@{}c@{}}72.7 $\pm$ 1.9\\0.47 MB\\\textbf{239.8 µs}\end{tabular} &
\begin{tabular}{@{}c@{}}69.0 $\pm$ 1.5\\\textbf{0.37 MB}\\907.8 µs\end{tabular} &
\begin{tabular}{@{}c@{}}61.1 (5.21)\\\textbf{0.64 (1.71)}\\ \textbf{239.8 (2.64)} \end{tabular} \\

\midrule

$\text{DFDT}_{\text{Medium}}$ &
\begin{tabular}{@{}c@{}}\textbf{75.1 $\pm$ 0.3}\\0.21 MB\\35.3 µs\end{tabular} &
\begin{tabular}{@{}c@{}}54.7 $\pm$ 0.5\\2.92 MB\\ \textbf{254.7 µs}\end{tabular} &
\begin{tabular}{@{}c@{}}79.4 $\pm$ 0.6\\0.10 MB\\39.0 µs\end{tabular} &
\begin{tabular}{@{}c@{}}\textbf{42.3 $\pm$ 0.5}\\0.47 MB\\112.3 µs\end{tabular} &
\begin{tabular}{@{}c@{}}\textbf{52.3 $\pm$ 0.6}\\0.20 MB\\69.4 µs\end{tabular} &
\begin{tabular}{@{}c@{}}70.9 $\pm$ 1.3\\1.12 MB\\260.5 µs\end{tabular} &
\begin{tabular}{@{}c@{}}72.9 $\pm$ 5.4\\1.23 MB\\942.8 µs\end{tabular} &
\begin{tabular}{@{}c@{}}63.9 (2.43)\\0.89 (3.71)\\246.4 (4.14)\end{tabular} \\

\midrule

$\text{DFDT}_{\text{High}}$ &
\begin{tabular}{@{}c@{}}74.6 $\pm$ 0.5\\0.21 MB\\34.0 µs\end{tabular} &
\begin{tabular}{@{}c@{}}\textbf{54.8 $\pm$ 0.4}\\12.19 MB\\280.5 µs\end{tabular} &
\begin{tabular}{@{}c@{}}79.0 $\pm$ 0.4\\ 0.05 MB \\ 36.2 µs\end{tabular} &
\begin{tabular}{@{}c@{}}41.7 $\pm$ 0.6\\3.21 MB\\105.7 µs\end{tabular} &
\begin{tabular}{@{}c@{}}51.8 $\pm$ 1.0\\0.20 MB\\70.3 µs\end{tabular} &
\begin{tabular}{@{}c@{}}\textbf{80.1 $\pm$ 1.1}\\1.05 MB\\255.3 µs\end{tabular} &
\begin{tabular}{@{}c@{}}\textbf{84.7 $\pm$ 2.7}\\1.23 MB\\942.9 µs\end{tabular} &
\begin{tabular}{@{}c@{}}\textbf{66.7 (2.14)}\\2.59 (4.14)\\244.9 (3.42)\end{tabular} \\

\bottomrule
\end{tabular}
\caption{Prequential accuracy (\%), memory usage (MB), and per-instance computational time (µs)}
\label{tab:merged_results}
\end{table*}

The ranking for memory consumption exhibits an inverse trend compared to accuracy. VFDT is the most memory-intensive algorithm, receiving the worst average rank. In contrast, $\text{DFDT}_{\text{Low}}$ achieves the best overall rank. Notably, VFDT performs significantly worse than both $\text{DFDT}_{\text{Low}}$ and SVFDT, whereas $\text{DFDT}_{\text{Medium}}$ and $\text{DFDT}_{\text{High}}$ do not differ significantly, highlighting their well rounded performance.

The CDD for runtime identifies VFDT as the slowest algorithm. Among the compared methods, only SVFDT demonstrated statistically significant better performance than VFDT. Although the variant $\text{DFDT}_{\text{Low}}$ exhibited a lower mean runtime than SVFDT, its ranking was not consistently superior across datasets.

\section{Conclusions}
\label{sec:conclusions}

Dynamic Fast Decision Tree (DFDT) combines different adaptive strategies from prior research to control decision tree growth. These features make DFDT particularly suitable for resource-constrained environments, such as IoT devices, edge computing applications, and real-time systems.

Extensive experimental evaluation, including ablation studies and statistical comparisons across a diverse set of data streams, demonstrated that DFDT variants consistently outperform or match baseline learners. In particular, $\text{DFDT}_{\text{Low}}$ was identified as the most resource-efficient configuration, achieving strong performance with minimal resource consumption. $\text{DFDT}_{\text{Medium}}$ offered a favorable trade-off between predictive accuracy and efficiency, while $\text{DFDT}_{\text{High}}$ achieved the highest accuracy overall, justifying increased memory and runtime costs when accuracy is the primary concern.

Unlike traditional VFDT-based learners, DFDT incorporates fine-grained control over growth and memory usage without sacrificing model adaptability. Moreover, its compatibility with existing ensemble frameworks allows it to serve as a drop-in replacement for VFDT, offering immediate benefits in large-scale, real-time stream mining applications. Future work may explore the integration of DFDT in different ensemble frameworks and investigate its performance on a broader range of datasets, including noisy imbalanced high-dimensional data streams.

\section{Acknowledgments}

Work funded by Portuguese Foundation for Science and Technology under Ph.D. scholarship PRT/BD/154713/2023 and project doi.org/10.54499/UIDP/00760/2020.

\bibliography{aaai2026}

\end{document}